\documentclass[runningheads]{llncs}
\usepackage[T1]{fontenc}
\usepackage{graphicx}
\usepackage{booktabs}
\usepackage[misc]{ifsym}

\usepackage{mwe}
\usepackage{amsmath}
\usepackage{amssymb}
\usepackage{booktabs}
\usepackage{subcaption}
\usepackage{array}  
\usepackage{hyperref}

\usepackage{todonotes}

\begin{document}

\title{Masked Autoencoder Self Pre-Training 
for Defect Detection in Microelectronics}

\titlerunning{MAE Self Pre-Training for Microelectronics}
% If the full title of your paper is short enough to also fit in the running head, you can omit the abbreviated paper title here. You can check as follows: if you comment out the \titlerunning line, something will appear in the header of all odd-numbered pages of your PDF from page 3 onward. This something is either the full title (in which case all is well), or the error message "Title Suppressed Due to Excessive Length". If this error message appears, you're going to want to provide an abbreviated title within the \titlerunning command, because if you won't do it, Springer will do it for you.

%N.B.: Author information (both in the \author{} and \authorrunning{} command) should only be present in the Camera-Ready Version of your paper. The version that you initially submit for review, ought to be double-blind. So, when initially submitting your paper, use:
%\author{Author information scrubbed for double-blind reviewing.}
\author{Nikolai Röhrich\inst{1,2} \and
Alwin Hoffmann\inst{1} \and
Richard Nordsieck\inst{1} \and
Emilio Zarbali\inst{1} \and 
Alireza Javanmardi \inst{2,3}}
% You may leave out the orcidID information, if you want to.
% Use \corr to indicate the corresponding author. Note the spacing around the \corr command. Only one author can be the corresponding author.

%N.B.: comment out the \authorrunning{} command for the double-blind version of your paper submitted for review. Later, if your paper is accepted, use the command for the Camera-Ready Version.
\authorrunning{Röhrich et al.}
% First names are abbreviated in the running head.
% If there is one author, write 'A.L. Benjamin'.
% If there are two authors, write 'A.L. Benjamin and C.C. Broadus Jr.'
% If there are more than two authors, '[...] et al.' is used.

\institute{XITASO GmbH, Germany
\and
Institute of Informatics, LMU Munich, Germany 
\and
Munich Center for Machine Learning (MCML), Germany
}

\maketitle              % typeset the header of the contribution

\begin{abstract}

While transformers have surpassed convolutional neural networks (CNNs) in various computer vision tasks, microelectronics defect detection still largely relies on CNNs. We hypothesize that this gap is due to the fact that a) transformers have an increased need for data and b) (labelled) image generation procedures for microelectronics are costly, and data is therefore sparse. Whereas in other domains, pre-training on large natural image datasets can mitigate this problem, in microelectronics transfer learning is hindered due to the dissimilarity of domain data and natural images. We address this challenge through self pre-training, where models are pre-trained directly on the target dataset, rather than another dataset. We propose a resource-efficient vision transformer (ViT) pre-training framework for defect detection in microelectronics based on masked autoencoders (MAE). We perform pre-training and defect detection using a dataset of less than 10,000 scanning acoustic microscopy (SAM) images. Our experimental results show that our approach leads to substantial performance gains compared to a) supervised ViT, b) ViT pre-trained on natural image datasets, and c) state-of-the-art CNN-based defect detection models used in microelectronics. Additionally, interpretability analysis reveals that our self pre-trained models attend to defect-relevant features such as cracks in the solder material, while baseline models often attend to spurious patterns. This shows that our approach yields defect-specific feature representations, resulting in more interpretable and generalizable transformer models for this data-sparse domain.

\end{abstract}

\section{Introduction}

\begin{figure}[t]
\includegraphics[width=\textwidth]{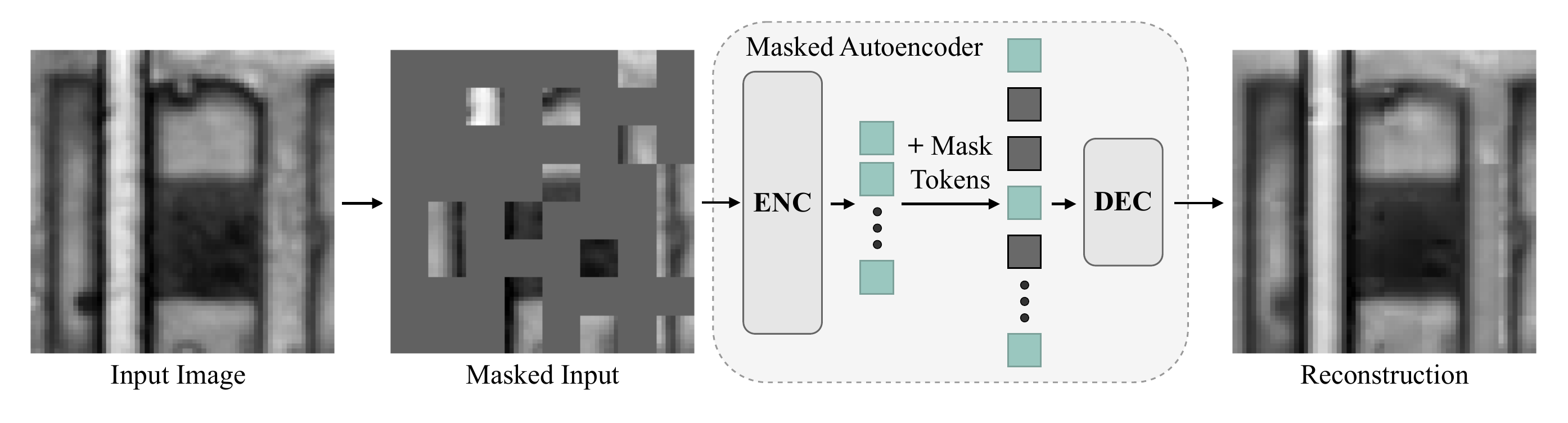}
\caption{\textbf{Masked Autoencoder for Microelectronics.} In MAE pre-training, a large share of input patches is masked. The encoder (ENC) then performs self-attention for visible patches only. After adding mask-tokens, which serve as placeholders for missing inputs, the decoder (DEC) reconstructs the complete image. Thereby, the model learns meaningful representations of input data.} \label{fig_mae}
\end{figure}

Reliable solder joints are crucial for the continued miniaturization and enhanced functionality of microelectronics, with applications spanning consumer electronics, automotive systems, healthcare, and defense \cite{njoku2023effects}. Defect detection in microelectronics often relies on images, which are obtained by intricate procedures like scanning microscopy or X-ray imaging. Image data are then used in smart manufacturing processes such as Automated Optical Inspection (AOI) to ensure product reliability and performance \cite{abd2020review,Schmid2022}. 

% In particular, deep learning based computer vision models have shown strong performance in AOI tasks like defect detection  \cite{villalba2019deep,wang2020remaining,saberironaghi2023defect,islam2024deep}. 

\begin{figure}[t]
    \centering
    % First subfigure (full width, first row)
    \begin{subfigure}{0.9\textwidth}
        \centering
        \includegraphics[width=\textwidth]{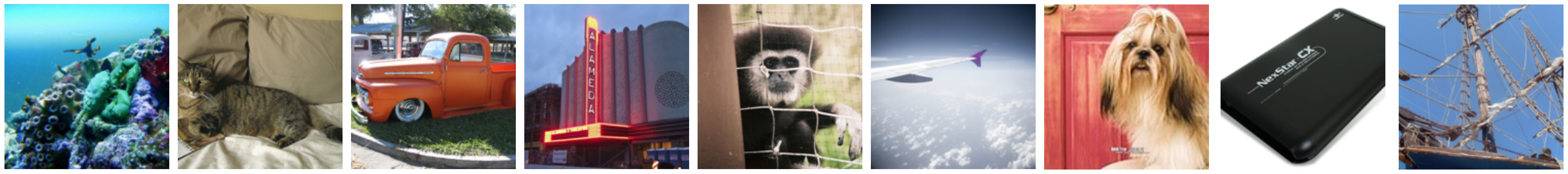}
        \caption{Natural image data from ImageNet \cite{deng2009imagenet}.}
        \label{fig:sub1}
    \end{subfigure}
    
    \vspace{0.1cm} % Space between rows
    
    % Second subfigure (full width, second row)
    \begin{subfigure}{0.9\textwidth}
        \centering
        \includegraphics[width=\textwidth]{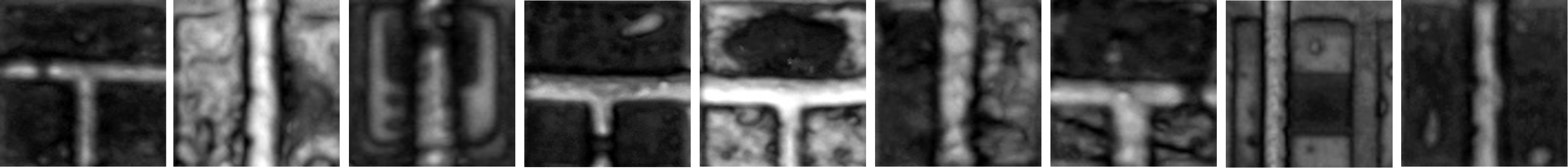}
        \caption{Scanning Acoustic Microscopy images from our target dataset \cite{Schmid2022}.}
        \label{fig:sub2}
    \end{subfigure}
    
    \caption{Domain gap between natural images and images from our target domain.}
    \label{fig_imagenet_vs_sam}
\end{figure}

Recently, vision transformer (ViT) models have emerged as the gold standard in computer vision \cite{khan2023survey,mauricio2023comparing}. However, the application of transformer models to defect detection in microelectronics remains underexplored, with the field largely relying on convolutional neural networks (CNN) and recent survey studies sometimes not mentioning transformer models at all \cite{saberironaghi2023defect,islam2024deep}. One main reason for this could be that training transformers from scratch usually requires a large amount of data \cite{Vaswani2017,Dosovitskiy2020,zhai2022scaling}. In microelectronics manufacturing - especially for microscale solder joints - labelled image data collection often relies on expensive procedures, resulting in small and imbalanced datasets \cite{abd2020review}. Additionally, fine-tuning transformers pre-trained on large datasets usually requires at least some similarity between the pre-training domain and the target domain \cite{Alijani2024}. This poses a challenge for microelectronics since most pre-trained transformers are trained on natural image datasets like ImageNet \cite{deng2009imagenet} that are highly dissimilar from target datasets in this domain (see Figure~\ref{fig_imagenet_vs_sam}).

Inspired by other domains facing similar challenges, such as healthcare, this paper explores the potential of a pure transformer model for microelectronics defect detection by leveraging self pre-training, where models are pre-trained directly on the target dataset rather than an extensive natural image dataset \cite{Zhou2023}. Specifically, our framework is based on masked autoencoders (MAEs) \cite{He2022}, which mask a large share of image patches and task the model with reconstructing the missing inputs (see Figure \ref{fig_mae}).
MAEs are resource-efficient and well-suited for smaller datasets, as they do not require large batch sizes like contrastive pre-training approaches. Additionally, the repeated randomized masking of different image patches presents the autoencoder with diverse reconstruction tasks, even with limited data. Using a dataset of less than 10,000 scanning acoustic microscopy (SAM) images of microscale LED solder joints labelled using transient thermal analysis (TTA), we compare our approach to a) purely supervised ViT, b) ViT pre-trained on a natural image dataset, and c) state-of-the-art CNN architectures proposed in the literature on industrial defect detection. In particular, we train our models for a regression task where models predict how far a given LED is from failure. We make the following contributions:

%\subsubsection{Contributions} 

\begin{itemize}
    \item We adapt masked autoencoder pre-training to defect detection under severe data scarcity in the microelectronics domain by self pre-training models directly on the target dataset, rather than on large natural image datasets.
    \item We demonstrate that this domain-specific self pre-training on less than 10,000 SAM images significantly outperforms supervised ViTs, ImageNet-pretrained ViTs, and several CNN architectures for defect detection. Our largest model improves mean squared error by up to 25\%, while requiring less than 12 hours of GPU time on a single A100 for both pre-training and fine-tuning, enabling resource-efficient application of vision transformers in this data-scarce domain.
    \item We show that, compared to baseline models, our approach yields defect-specific feature representations in pure transformer architectures. Our self pre-trained models focus on meaningful defect regions, leading to improved interpretability and generalizability, whereas baseline approaches often attend to irrelevant features or learn shortcuts.
\end{itemize}

%Our experimental findings indicate that, with our approach, models learn generalizable representations in self pre-training that yield substantial performance increases in defect detection for microelectronics. In particular, our model outperforms a) purely supervised ViTs by $21.1 \%$, b) ViTs pre-trained on Imagenet by $10.2 \%$ and c) state-of-the-art CNN defect detection models by $8.9 \%$. Also, we find that our models learn fault-specific representations, focusing on actual damages in the solder material when performing defect detection, while other ViT models have scattered attention or learn shortcuts. In addition, due to the simplicity and low data requirements of our framework, it can be easily integrated into real-world inline quality control procedures such as AOI. 

%----------------------------------------------------------

\section{Related Work}

\subsubsection{Vision Transformer} ViT \cite{Dosovitskiy2020}  is a vision adaption of the self-attention based transformer architecture \cite{Vaswani2017}. Related research on ViT includes domain-specific fine-tuning of foundational models and architectural adaptions that aim at improved domain generalization or increased data efficiency: Alijiani et al. \cite{Alijani2024} discuss various ViT domain adaption and generalization approaches on the feature extraction level, the instance level and the model architecture level. Since we leverage representations learned in pre-training, our approach is similar to domain specific fine-tuning. However, our framework differs from such approaches since we pre-train ViT from scratch on the target dataset, rather than using foundational models pre-trained on large natural image datasets. 

\subsubsection{Self-Supervised Learning} SSL aims to learn meaningful representations from data without labels. Instead of labels, intrinsic relationships of the data serve as guidance for the training process. Balestriero et al. \cite{Balestriero2023} distinguish four sub-types of SSL approaches: \textit{(1)} deep metric learning approaches such as SimCLR \cite{Chen2020-1} or NNCLR \cite{Dwibedi2021}, where models recognize similarity in differently transformed inputs, \textit{(2)} self-distillation learning approaches like BYOL \cite{Grill2020}, SimSIAM \cite{Chen2020} or DINO \cite{Caron2021}, where different input transforms serve as inputs into two separate encoders before letting one serve as a predictor of the other's output, \textit{(3)} canonical correlation analysis approaches such as VICReg \cite{Bardes2021}, BarlowTwins \cite{Zbontar2021} or SWAV \cite{Caron2020}, which are founded on the principle of understanding relationships between two variables by examining their cross-covariance matrices and \textit{(4)} masked image modelling (MIM) approaches like SimMIM \cite{Xie2021}, where images undergo partial masking and models reconstruct the missing inputs. We follow the MIM approach due to its its simplicity and effectiveness in low-data environments.

In MIM, iGPT \cite{Chen2020-2} first demonstrated the potential of applying masked language modeling strategies to the vision domain by predicting masked pixels in a sequence-like manner. BEiT \cite{Bao2021} introduced a discrete token prediction objective analogous to masked language modeling in NLP and showed that visual tokens enhance pre-training for image-related tasks. In MAE \cite{He2022}, an encoder receives approximately $25\%$ of the unmasked patches as input, and a lightweight decoder reconstructs the full image (see Figure \ref{fig_mae}). MAE pre-training has been demonstrated to achieve state-of-the-art performance with minimal data and compute requirements \cite{zhang2023survey}. Our approach builds on the work of He et al. \cite{He2022}, adapting their framework to the requirements of our use case.

\subsubsection{Microelectronics Quality Control} Deep learning based approaches have significantly advanced the field in recent years, including but not limited to the subdomain of solder joints \cite{saberironaghi2023defect,islam2024deep}. Samavatian et al. \cite{Samavatian2021} present an iterative machine learning framework that enhances the accuracy of solder joint lifetime prediction. This is achieved by utilizing a self-healing dataset that is iteratively injected into a correlation-driven neural network (CDNN). Salameh et al. \cite{Salameh2022} demonstrate the use of deep neural networks, combined with finite element simulations, as a rapid and comprehensive tool for solder joint reliability analysis under mechanical loading. Muench et al. \cite{Muench2022} propose a methodology for predicting damage progression in solder contacts and compare a multi-layer perceptron network with a long short-term memory (LSTM) model using production-like synthetic data. Zhang et al. \cite{Zhang2022-DeepLSolderJoint} employ various deep learning models, including CNN and LSTM models, for automatic solder joint defect detection using X-ray images. Zhang et al. \cite{Zhang2023-SolderJointQuality} combine CNN and transformer components in a model intended for printed control board solder joints. However, Zhang et al. use transformer blocks only as small parts of their network aided by convolutional layers and report poor performances for pure transformer models \cite{Zhang2023-SolderJointQuality}. We believe that this is due to a lack of domain-specific pre-training. Our approach, in contrast, demonstrates the potential of pure transformer architectures for defect detection without architectural crutches like CNN layers.

%----------------------------------------------------------
\section{Dataset}

We use scanning acoustic microscopy (SAM) images and transient thermal analysis (TTA) measurements from a dataset of microscale solder joints of high-power LEDs~\cite{Schmid2022}, which is publicly available on Kaggle~\cite{hella_kaggle_2023}. The dataset contains SAM and TTA data for 1,800 LEDs at 5 points of time, with nine LED types and five lead-free solder pastes. To simulate ageing, LED panels underwent thermal shock cycles (TSC) from $-40$\textdegree C to $125$\textdegree C, a standard quality control process to evaluate the reliability and durability of solder joints under extreme conditions. We use a train, validation and test split of 60\%, 20\% and 20\% respectively. All reported metrics and all shown example outputs refer to the test dataset.

\begin{table}[t]
  \centering
  \caption{\textbf{Dataset.} Repeated cooling and heating of LED solder joints simulates aging through thermomechanical fatigue. Over time, cracks gradually emerge, impeding the heat flow through the solder material. $\Delta B_{max}$ values indicate relative degradation and LEDs with $\Delta B_{max} > 20\%$ are classified as defective $(*)$. }
  \vspace{2mm}
  \begin{tabular}{b{1.25cm}  *{5}{c}}
    \textbf{TSC} & \textbf{0} & \textbf{100} & \textbf{500} & \textbf{1,000} & \textbf{1,500} \\
    \toprule
    
    &
    
    \includegraphics[width=0.15\textwidth]
    {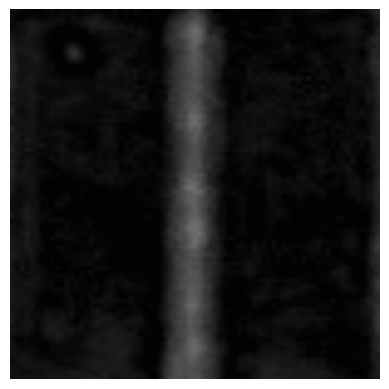}
    & 
    \includegraphics[width=0.15\textwidth]{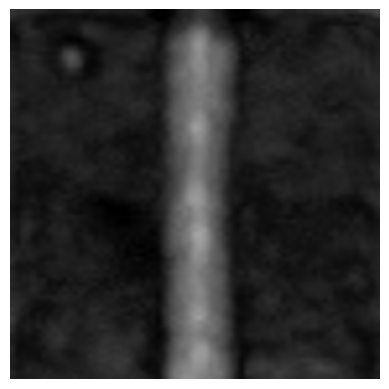} 
    & 
    \includegraphics[width=0.15\textwidth]{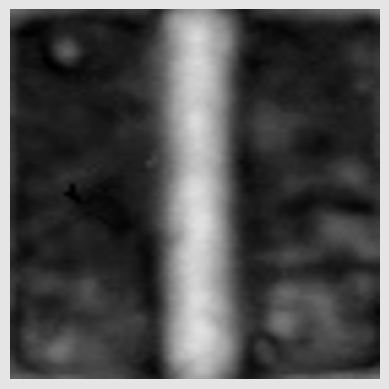}
    & 
    \includegraphics[width=0.15\textwidth]{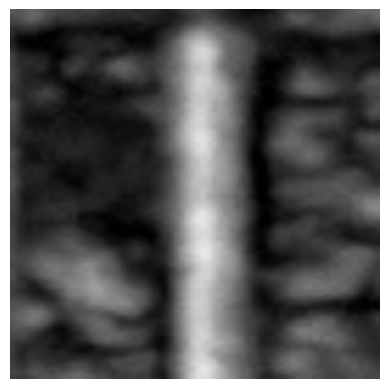}
    & 
    \includegraphics[width=0.15\textwidth]{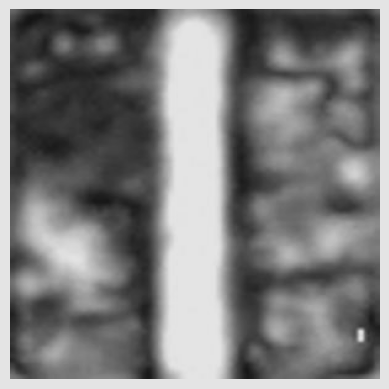}
    \\ 
    $\Delta B_{max}$ & $0\%$ & $3\%$ & $18\%$ & $99\% $ $(*)$ & $106\%$ $(*)$ \\
    \toprule

    & 
     
    \includegraphics[width=0.15\textwidth]{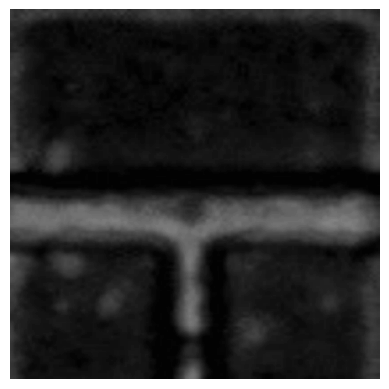} 

    & 
    \includegraphics[width=0.15\textwidth]{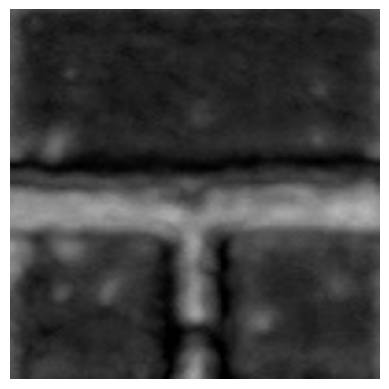} & 
    \includegraphics[width=0.15\textwidth]{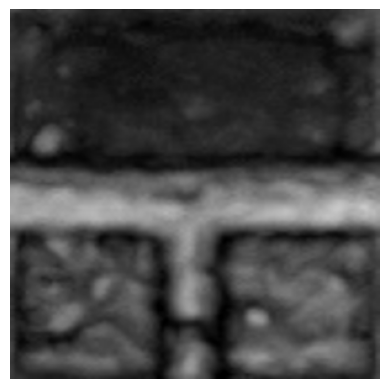} 
    & 
    \includegraphics[width=0.15\textwidth]{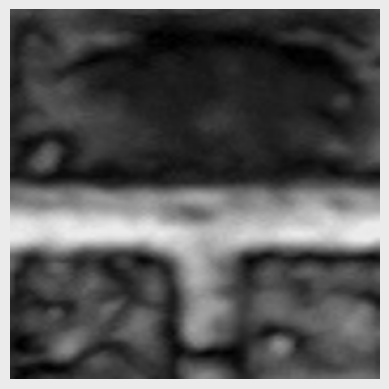}
    & 
    \includegraphics[width=0.15\textwidth]{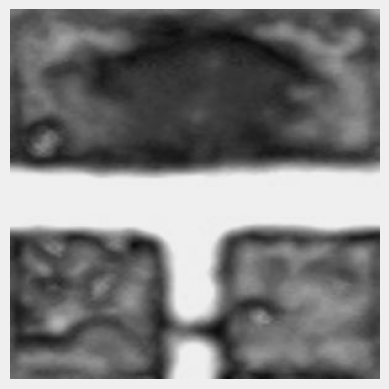}
    \\
    $\Delta B_{max}$ & $0\%$ & $2\%$ & $30\%$ $(*)$& $62\% $ $(*)$ & $75\%$ $(*)$ \\
    \toprule
  \end{tabular}
  
    \label{fig:five_images}
\end{table}

\subsection{Scanning Acoustic Microscopy (SAM)} 
We use SAM images for pre-training and as inputs for defect detection. SAM imaging offers detailed information on material interfaces, allowing the detection of critical defects such as voids, cracks, and delaminations within solder joints \cite{Yazdan2015}. SAM images were taken after 0, 100, 500, 1,000, and 1,500 TSC, resulting in 5 images per LED and 9,000 images in total (see Table \ref{fig:five_images}). The original SAM images contain 2 LEDs and are cropped so that each image shows an individual solder joint in a $64 \times 64 \times 1$ pixel format. For each LED image, bright boundary regions depict gaps between the darker solder pads. On the solder pads, small circles indicate voids in the solder material, from which cracks can emerge over time. Bright, non-circular structures inside the pads are cracks in the solder material (see Table \ref{fig:five_images}). 

\subsection{Transient Thermal Analysis (TTA)}
We use TTA data to create labels indicating how far LEDs depicted in SAM images are from failure. TTA is a non-destructive test method that quantifies the integrity of the thermal path from the LED junction to the heat sink through the solder joint~\cite{zippelius_lstm}. Defects such as cracks impede the heat flow through the solder by reducing the size of the contact area. In TTA, the thermal impedance $Z_{th}(t)$ is measured, which describes the temperature difference at the LED junction over time compared to the change in power. Any degradation results in an increase in $Z_{th}(t)$. For a quantitative comparison of TTA measurements between different TSCs, the maximum value $B_{max}$ of the normalized logarithmic derivative $B(t')$ of the thermal impedance $Z_{th}(t)$ with $t' = ln(t)$ is used~\cite{SAM-CNN}. As $B_{max}$ increases in case of degradation, the relative increase in $B_{max}$ at cycle $t$ compared to the initial state $\Delta B_{max}(t)$ serves as the basis for image labelling: 

\begin{equation}
    \Delta B_{max}(t) = \frac{B_{max}(t)}{B_{max}(0)}-1.
\end{equation}

\noindent While we use numerical labels to achieve a continuous representation of an LED's current quality, an LED with $\Delta B_{max} > 20\%$ is classified as defective \cite{Zarbali2023}.

%----------------------------------------------------------
\section{Method}

\subsection{Preliminaries}

\subsubsection{Vision Transformer} Transformer processing employs multi-head self-attention (MSA) on sequential inputs \cite{Vaswani2017}. For the sequentialization of 2D images, images are divided into smaller patches and are linearly projected into the embedding space. Let $H_i, W_i$ be the height and width of the original image, and let $H_p, W_p$ be the patch height and width, where usually image patches are quadratic, i.e. $H_p = W_p =: P$ (\textbf{$*$}). The image is then divided into

\begin{eqnarray}
    N = \frac{H_i \cdot W_i}{H_p \cdot W_p}  \overset{\textbf{$*$}}{=}  \frac{H_i \cdot W_i}{P^2}
\end{eqnarray}

\noindent patches. After flattening and applying a linear mapping $E \in \mathbb{R} ^ {P^2 \times D}$ to project the patch into the embedding space $\mathbb{R}^D$, positional embeddings $E_{pos} \in \mathbb{R}^{N \times D}$ are added to the elements of the sequence to retain the spatial location of each patch embedding. Lastly, a learnable class token $x_{class}$ is prepended to the sequence to store global information relevant to the $\Delta B_{max}$ prediction during processing. Images are thus transformed into a sequence $X$ of $N+1$ embeddings, which serve as tokens for MSA:

\begin{eqnarray}
    X = [x_{class}, Ex_1 + E_{pos,1}, \dots, Ex_N + E_{pos,N}] \subseteq \mathbb{R}^D
\end{eqnarray}

Depending on the ViT size, the sequence passes through several MSA blocks. Since we are working with relatively small $64 \times 64 \times 1$ shaped images, we use ViT sizes tiny, small, and base for determining the number of transformer blocks, the token length, and the number of heads (see Table ~\ref{fig:vit_sizes}). In the original ViT \cite{Dosovitskiy2020}, the output class token is fed into a single linear layer for classification. We find that using a larger classification head increases performance and thus use a multi-layered dense network with a hidden dimension of $2048$. 

\begin{table}[t]
    \centering
    \caption{\textbf{ViT Sizes.} While all of the evaluated sizes use 12 transformer blocks, they differ in token length (width) and the number of heads used for MSA.}
    \vspace{2mm}
        \begin{tabular}{l@{\hspace{5mm}}c@{\hspace{5mm}}c@{\hspace{5mm}}c@{\hspace{5mm}}c}
             \hline
                Model       & Layers         & Width         & Heads         & Params \\ \hline
                ViT-Ti & 12 & 192 & 3 & 5.7 M  \\
                ViT-S & 12 & 384 & 6 & 22.1 M \\
                ViT-B & 12 & 768 & 12 & 86.7 M  \\
        \end{tabular}
    
    \label{fig:vit_sizes}
\end{table}

\subsubsection{Masked Autoencoder Pre-Training} For MAE pre-training, a ViT encoder combined with a lightweight transformer decoder is tasked with reconstructing missing image patches (see Figure~\ref{fig_mae}). Depending on the masking share $S$, which is usually  $0.75$, a random subset $M$ of $\lfloor  N \cdot S \rfloor$ indexes is sampled from the set of indexes $I = \{1, \dots, N\}$ without replacement. $X$ is split into a set of masked patches $X_M$ and a set of visible patches $X_{I \setminus M}$. Note that $X_M$ is the absolute complement of $X_{I \setminus M}$ with respect to $X_I$, i.e. $X_m \cup X_{I \setminus M} = X_I$ and $X_m \cap X_{I \setminus M} = \emptyset$. After masking, the sequence consists of $\lfloor  N \cdot S \rfloor$ embeddings:

\begin{eqnarray}
    X = [ \Tilde{x}_1E + E_{pos,1}, \dots, \Tilde{x}_{\lfloor  N \cdot S \rfloor}E + E_{pos,\lfloor  N \cdot S \rfloor}]  \subseteq \mathbb{R}^D
\end{eqnarray}

Inputs are then fed into the autoencoder network. The MAE encoder operates only on the visible patches, i.e. on a $1-S$ share of the total input patches. The decoder, in contrast, receives as input a full sequence of tokens including a) the embeddings of \textit{visible patches} and b) \textit{mask tokens}, which are shared and learnable embeddings serving as placeholders for masked patches, similar to BERT \cite{Devlin2018}. The model minimizes the pixel-wise mean squared error (MSE) loss for masked patches with respect to its parameters $\theta$. The loss is computed for masked patches only, i.e. for $k \in M$: 

\begin{eqnarray}
    \min_{\theta} \text{ } \mathcal{L}_{mae}(X, M, \theta) = \sum_{k \in M} {\|y_k - x_k\|_2}^2
\end{eqnarray}

\begin{table}[t]
    \centering
    \caption{Hyperparameter tuning for patch size, mask ratio, and augmentations.}
    \begin{subtable}{0.3\textwidth}
        \centering
        \caption{Patch Sizes}

        \begin{tabular}{c@{\hspace{5mm}}c}
            \toprule
            Patch Size  	          & MSE $\downarrow$   \\  
            
            \bottomrule

            $4 \times 4$               & 0.0186  	 \\ 
            \textbf{8 $\times$ 8} & 	\textbf{0.0167} \\ 
            $16 \times 16$		      & 0.0197 \\
            \bottomrule
        \end{tabular}
        \label{tab:sub1}
    \end{subtable}
    \hspace{2mm}  % Reduce space instead of \hfill
    \begin{subtable}{0.3\textwidth}
        \centering
                \caption{Mask Ratios}

        \begin{tabular}{c@{\hspace{5mm}}c}
            \toprule
            Mask Ratio	 & MSE $\downarrow$  	 \\ 
            \bottomrule
            70\%		     & 0.0175
               \\ 
            \textbf{75\%}    & \textbf{0.0167}
	 \\ 
            80\%			 & 0.0188
        		 \\
            \bottomrule
        \end{tabular}
        \label{tab:sub2}
        \end{subtable}
        \hspace{2mm}  % Reduce space instead of \hfill
        \begin{subtable}{0.3\textwidth}
        \centering
                \caption{Augmentations}

        \begin{tabular}{c@{\hspace{5mm}}c}
        \toprule
        Augmentations 	 & MSE $\downarrow$  \\
        \bottomrule
        horizontal flip	     & 0.0190 \\ 
        vertical flip	         & 0.0189               \\
        \textbf{both}					 & \textbf{0.0179}         		 \\             \bottomrule

        crop (random)	        & 0.0185                  \\ 
        \textbf{crop (fixed)}	&  \textbf{0.0176}   \\ 
        \bottomrule
        \end{tabular}
        \label{tab:sub3}
    \end{subtable}
    \hspace{2mm}
    
    \label{tab:ablation}
\end{table}

\section{Experiments}

\subsection{Hyperparameter Tuning} 

Because of the distinct structure of SAM images and due to the smaller image size compared to standard datasets, we hypothesize that smaller patch sizes, different augmentations, and different mask ratios than those used for standard ViT could achieve optimal performance. To evaluate hyperparameter settings, we pre-train models with MAE for 200 epochs on our dataset, followed by supervised fine-tuning for defect detection over 100 epochs. We report the MSE with respect to the target $\Delta B_{\text{max}}$ after fine-tuning.

Results are shown in Table~\ref{tab:ablation}. Although we find that, consistent with the results for ImageNet, a mask ratio of $75 \%$ works best, our hypothesis was confirmed for patch size and augmentations: For image patches, a size of $8 \times 8$ worked best, while typically, a patch size of $16 \times 16$ is used. Notably, the found optimal image to patch size ratio of $64:8 = 8$ also differs from the standard ratio of $224:16 = 14$ used for datasets like ImageNet \cite{Dosovitskiy2020}. We assume that this non-linear dependence between image size and optimal patch size is explained by the fact that once patches get too small, a single patch does not contain enough localized information for a meaningful computation of self-attention. 

For augmentations, a combination of horizontal and vertical flipping and resized cropping using a fixed crop-size achieved the best results, while for ImageNet horizontal flipping and random-sized cropping are used \cite{Dosovitskiy2020}. In contrast to natural images, vertical flipping is a valid augmentation for microelectronics, where image orientation is not relevant. Also, we suspect that fixed-size cropping works best since random-sized crops of $64 \times 64$ images suffer greater quality loss by resizing operations than the $224 \times 224$ images in standard large datasets like ImageNet.

\begin{figure}
\centering
\includegraphics[width=0.75\textwidth]{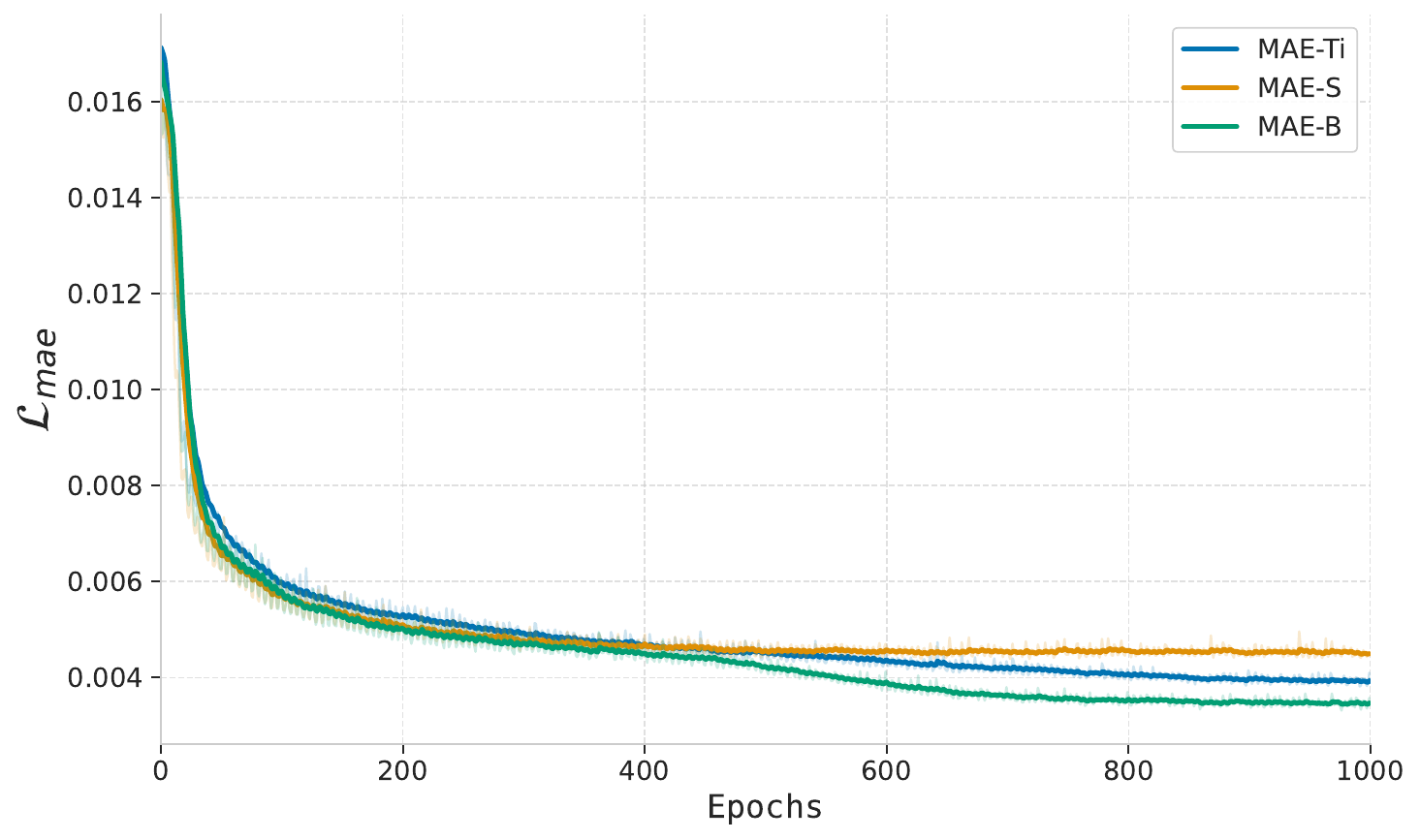}
\caption{\textbf{MAE Pre-Training.}} \label{fig_losses}
\end{figure}

\subsection{MAE Self Pre-Training}

We conduct MAE pre-training on the target SAM image dataset, utilizing the hyperparameters optimized through our previous  studies. MAE with ViT encoders of sizes Ti, S, and B are pre-trained for 1,000 epochs respectively. In Figure \ref{fig_losses}, we plot the MSE reconstruction loss for masked input patches $\mathcal{L}_{mae}$ for each model size. Notably, the S-sized encoder achieved the worst performance in reconstruction, although having more parameters than the Ti-sized encoder. However, the even larger B-sized encoder significantly outperformed all smaller sizes. We take this to indicate that a) for smaller images, even very small encoders like ViT-Ti can achieve good results and that b) there is a non-linear relationship between model capacity and performance. For smaller datasets like ours, there appears to be a critical threshold where increased model complexity transitions from potentially harmful (due to overfitting) to beneficial. Although ViT-B achieved the best results overall, ViT-Ti could be the model of choice in resource-critical applications. 

\begin{figure}
    \centering
    % First subfigure
    \begin{subfigure}{0.43\textwidth}
        \centering
        \includegraphics[width=\textwidth]{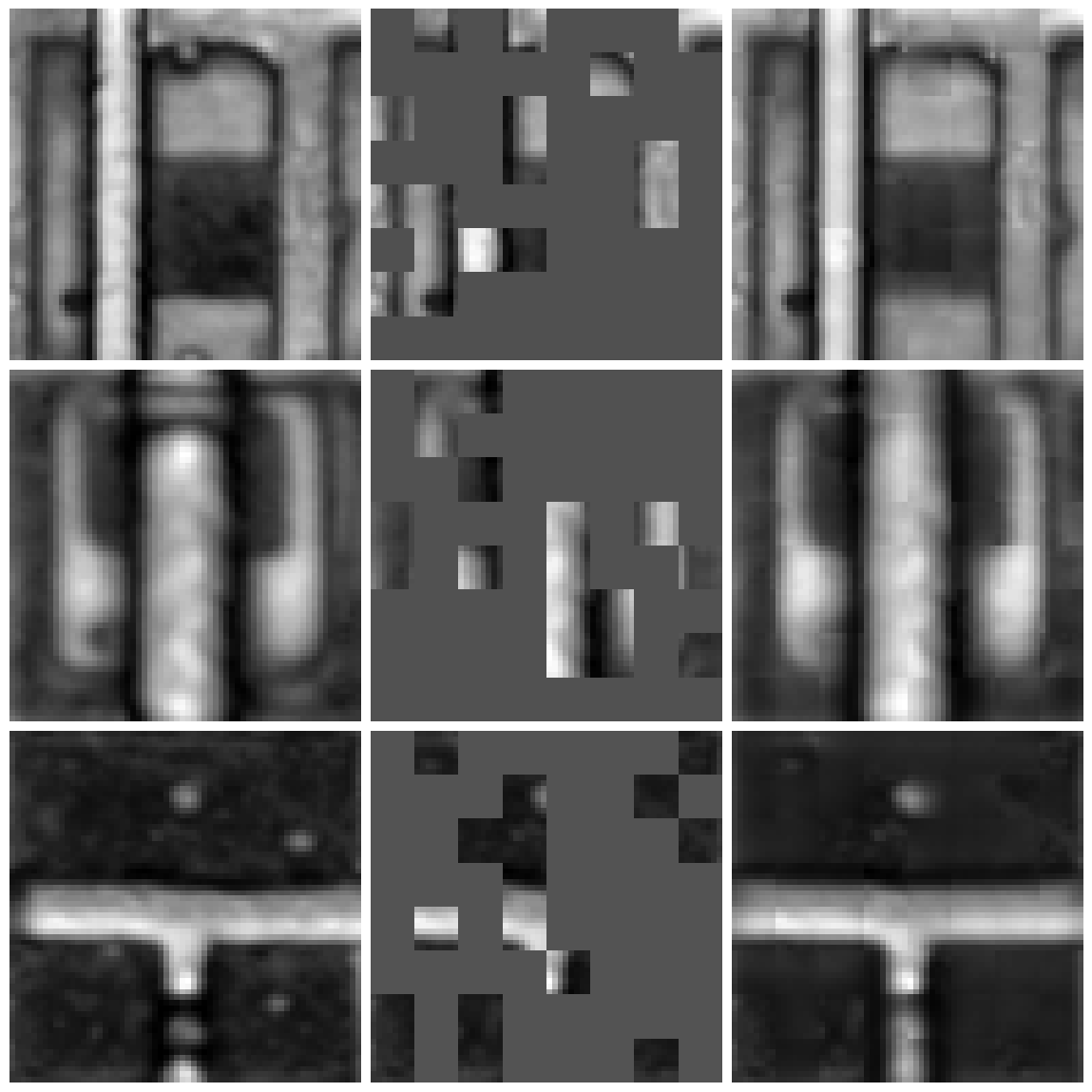} 
        \caption{\small 0 TSC}
        \label{fig:sub1}
    \end{subfigure}
    \hspace{2mm}
    % Second subfigure
    \begin{subfigure}{0.43\textwidth}
        \centering
        \includegraphics[width=\textwidth]{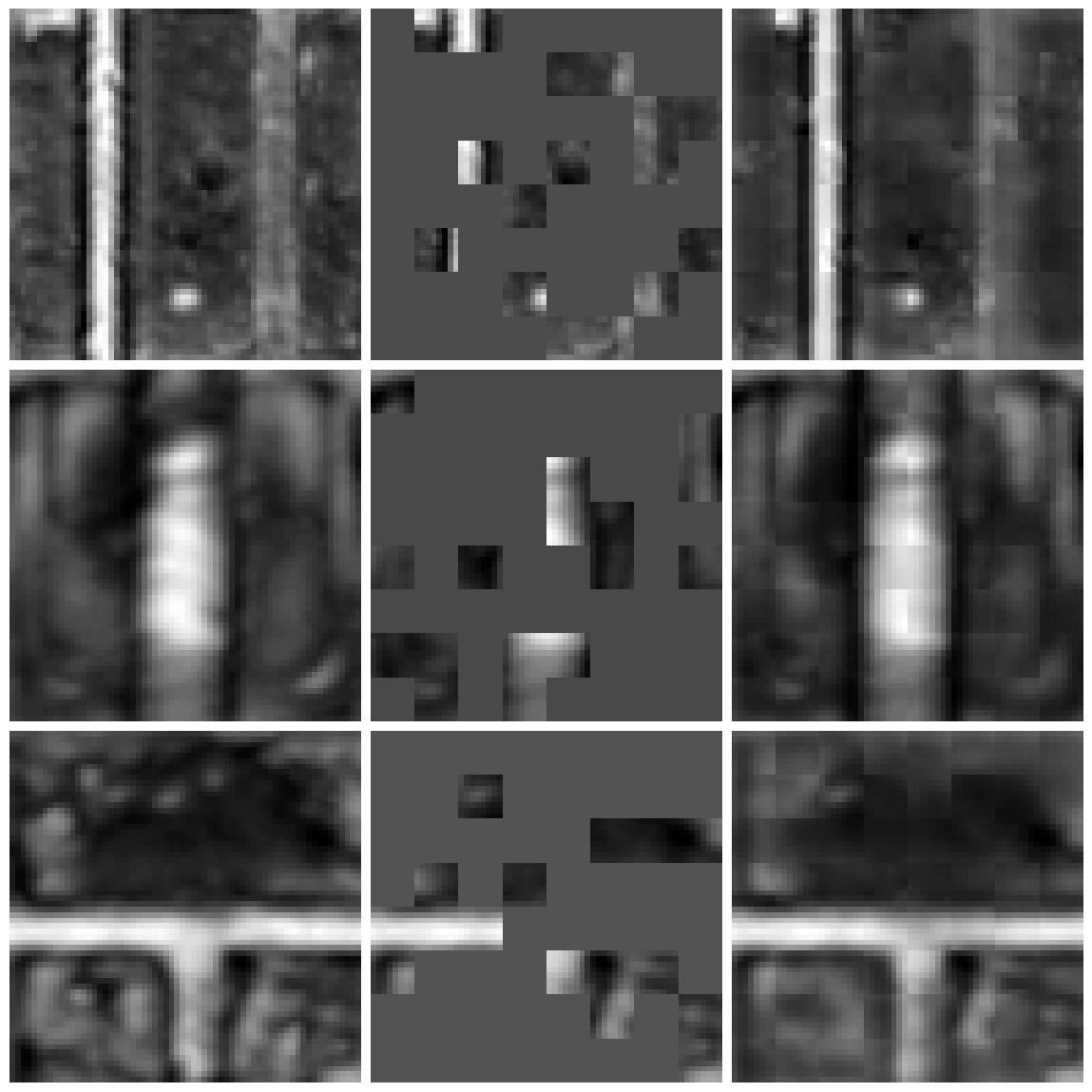} 
        \caption{\small 1,500 TSC}
        \label{fig:sub2}
    \end{subfigure}
    
    \caption{\textbf{MAE Reconstructions.} For (a) TSC 0 and (b) TSC 1,500, original images, masked inputs, and the model's reconstructions are depicted.}
    \label{fig:reconstructions}
\end{figure}

Exemplary reconstructions from the test set using the best-performing ViT-B encoder are shown in Figure \ref{fig:reconstructions}. We find that after training, the autoencoder produces high-quality reconstructions for all LED types, indicating that it learned meaningful representations of the dataset. In particular, the model is able to reconstruct masked inputs at varying amounts of TSC, accurately predicting the expansion of cracks even in masked regions (see Figure \ref{fig:reconstructions} (b)).

\textbf{Computational Effort} All MAE pre-training experiments were conducted on a single A100 GPU. For our best-performing model ViT-B, the full pre-training process completes in under 8 hours, and fine-tuning for defect detection takes less than 1 hour. For the ViT-Ti model, which already outperforms all evaluated baselines on defect detection, the whole process of pre-training and fine-tuning takes under 2 hours. This highlights the computational accessibility of our approach, even in constrained environments.

\subsection{Defect Detection}                                                          

For inline automated optical inspection in microelectronics manufacturing, defect detection is a common practice. We perform defect detection given an input image using the $\Delta B_{max}$ values given by TTA as labels. While an LED with $\Delta B_{max}$ larger than  $20\%$ is classified as defect, we predict the scalar $\Delta B_{max}$ values directly to have a continuous measure for LED quality. That is, having the model predict $\Delta B_{max}$ not only indicates whether the given LED is functional, but also how far the LED is from failure. Note that defect detection differs from anomaly detection where a pixel-wise labelling is performed (see e.g. \cite{chen2024unified}).

We compare ViT pre-trained with our MAE-based framework for 1,000 epochs against purely supervised ViT and ViT pre-trained with ImageNet. Also, we compare our models against large standard CNN backbones like ResNet \cite{ResNet} or EfficientNet \cite{EfficientNet} without additional pre-training, as well as against recent CNN architectures specifically designed for defect detection in microelectronics. In particular, we evaluate the model by Zippelius et al. \cite{SAM-CNN} intended for SAM images of solder joints as well as the model by Zhang et al. \cite{Zhang2022-DeepLSolderJoint} intended for X-ray images of solder joints. CNN models are trained in supervised fashion on defect detection without additional pre-training. All models are trained for up to 200 epochs.

We find that ViT pre-trained with our self pre-training framework outperforms all CNN and transformer-based models, achieving an MSE improvement of $8.9\%$ compared to the best-performing baseline (see Table~\ref{tab:fault_detection}). When it comes to other ViT models, our model outperforms purely supervised ViTs by $21.1 \%$ and ViTs pre-trained on Imagenet by $10.2 \%$. Notably, while the best results are achieved by the largest self pre-trained transformer model, ViT-Ti already outperforms considerably larger baselines by $6.9\%$ with only 5.7 million parameters. Thus, ViT self pre-trained with our approach not only substantially outperforms other ViT models and state-of-the-art CNN defect detection models, but is already able to do so using the smallest available architecture size.

\begin{table}
    \centering
    \caption{\textbf{Defect Detection.} All models are trained for 200 epochs on the SAM dataset. We report absolute MSE and relative difference compared to the best-performing model ($\mathrm{\Delta}$ to Best) to provide an intuitive sense of performance gains.}
    \vspace{1mm}
    \begin{tabular}{l@{\hspace{3mm}} c@{\hspace{3mm}} c@{\hspace{3mm}} c@{\hspace{3mm}} c}
        \toprule
        Model & Pre-Training & Params & MSE & $\mathrm{\Delta}$ to Best\\ 
        \midrule
        SAM-CNN \cite{SAM-CNN}& -  & 4.2 M & 0.0336 & 10.5\% \\
        XRAY-CNN \cite{Zhang2022-DeepLSolderJoint} & -  & 8.7 M & 0.0337 & 10.9\%\\
        ResNet50 \cite{ResNet} & -  & 23.5 M & 0.0339 & 11.5\%\\
        EfficientNet-B7 \cite{EfficientNet}& -  & 63.8 M & 0.0331& 8.9\% \\
        \midrule
        ViT-Ti \cite{Dosovitskiy2020}  & -  & 5.7 M & 0.0380 & 25.0\% \\
        ViT-S \cite{Dosovitskiy2020}  & - & 22.1 M & 0.0368 & 21.1\% \\
        ViT-B \cite{Dosovitskiy2020}  & -  & 86.7 M & 0.0380 & 25.0\% \\
        \midrule
        ViT-Ti \cite{Dosovitskiy2020} & MAE (ImageNet)  &  5.7 M & 0.0345 & 13.4\%\\
        ViT-S \cite{Dosovitskiy2020}  & MAE (ImageNet) &  22.1 M & 0.0339 & 11.5\%\\
        ViT-B \cite{Dosovitskiy2020}  & MAE (ImageNet)  & 86.7 M & 0.0335 & 10.2\%\\
        \midrule
        ViT-Ti (ours)  & MAE (self) & 5.7 M & 0.0310 & 2.0\% \\
        ViT-S (ours) & MAE (self) & 22.1 M & 0.0335 & 10.2\%\\
        ViT-B (ours)  & MAE (self) & 86.7 M & \textbf{0.0304} & 0.0\%  \\

        \bottomrule
    \end{tabular}
    
    \label{tab:fault_detection}
\end{table}         

In contrast to our self pre-trained models however, we find that purely supervised ViT performs poorly compared to all other models due to the lack of pre-training and the small amount of target data, which is consistent with earlier findings \cite{Zhang2023-SolderJointQuality}. What is more, we find that MAE pre-training using ImageNet increases performance compared to purely supervised ViT, but cannot outperform convolutional baselines. This suggests that for applications with even less data than in our case, fine-tuning foundational vision models or using convolutional defect detection architectures might remain a valid option.

\subsubsection{Interpretability Analysis} 
\begin{figure}[t]
\centering
    \begin{subfigure}{0.48\textwidth}
        \centering
        \includegraphics[width=\textwidth]{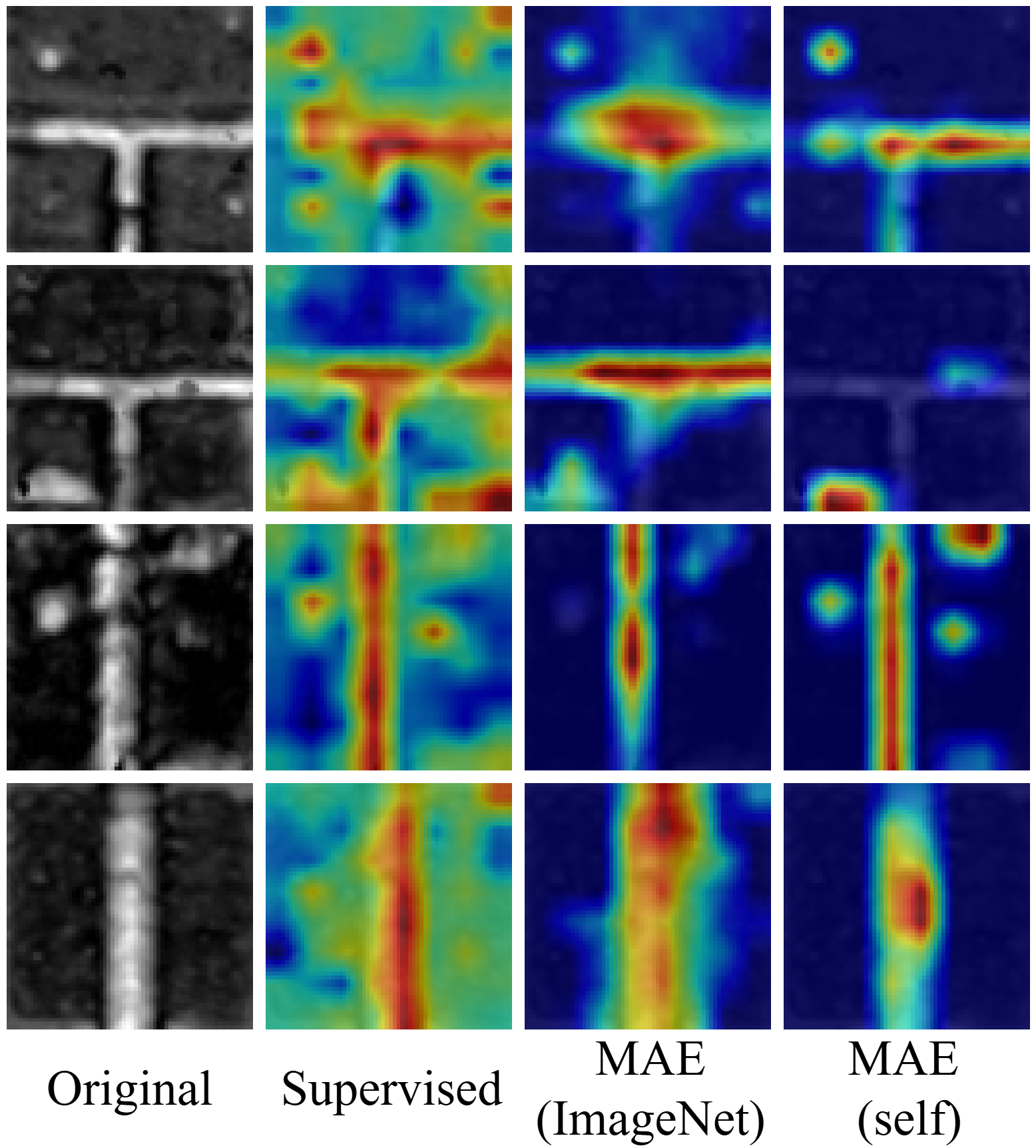} 
        \caption{\small 0 - 500 TSC}
        \label{fig:sub1}
    \end{subfigure}
    \hspace{2mm}
    % Second subfigure
    \begin{subfigure}{0.48\textwidth}
        \centering
        \includegraphics[width=\textwidth]{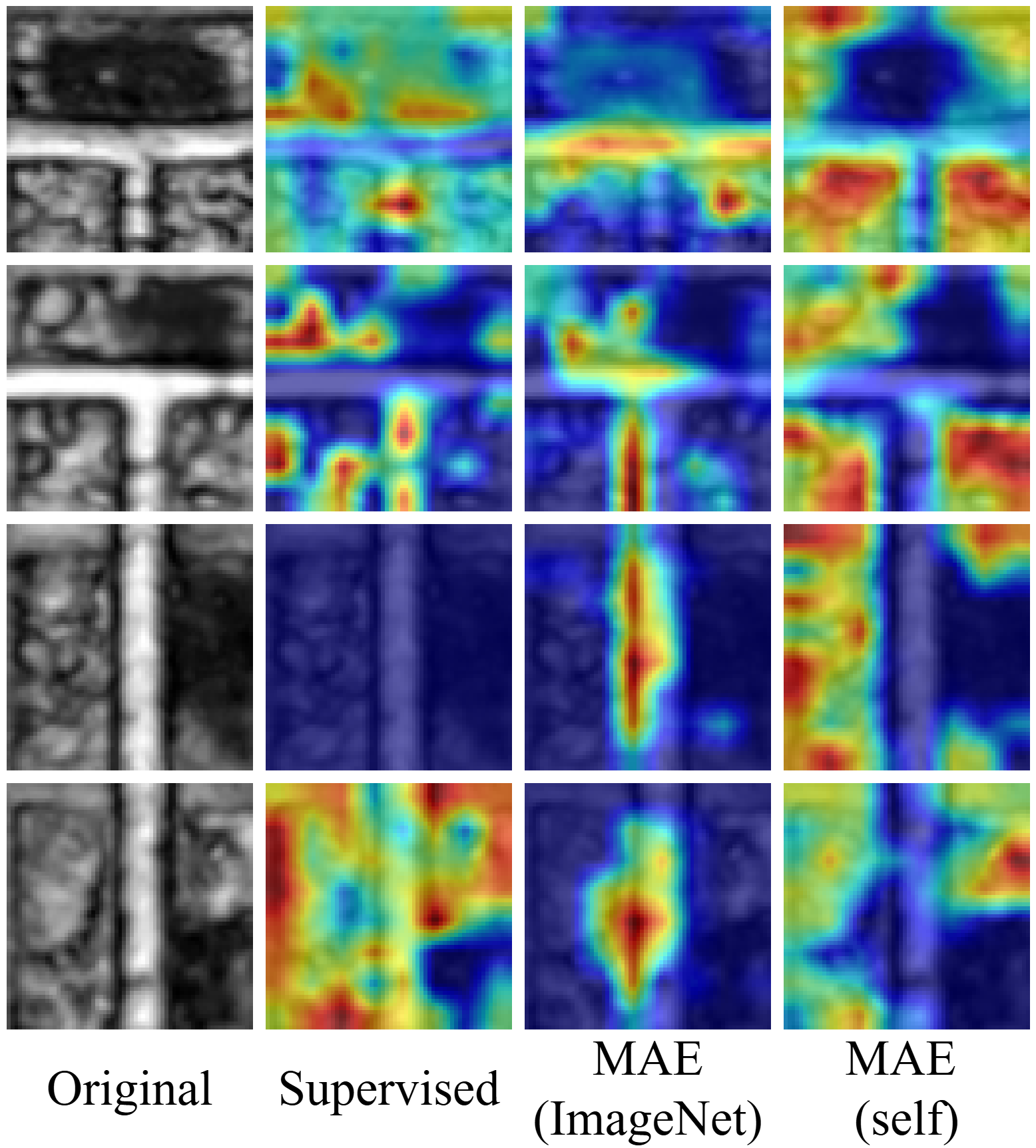} 
        \caption{\small 1,000 - 1,500 TSC}
        \label{fig:sub2}
    \end{subfigure}
\caption{\textbf{Attention Visualization.} GradCAM heatmaps for supervised ViT, ViT pre-trained on ImageNet, and our ViT self pre-trained on SAM data. Baseline models have scattered attention or learn shortcuts between LED type and defects, thus neglecting the true causal relationship between damages in the solder and LED defects. Our self-pre-trained model, in contrast, consistently focuses on general structural information when no defects are present (a), and pinpoints defect-relevant regions such as cracks whenever they are visible (b).} 
\label{fig:cradcam}
\end{figure}

We analyze class activation maps using GradCAM \cite{selvaraju2017grad}. In GradCAM, the relevance of image regions for the output is visualized by computing gradients of the output with respect to a given layer. We choose the first norm layer of the last transformer block for the gradient computation. Results for all ViT-B variants are shown in Figure~\ref{fig:cradcam}. We find that:
\begin{enumerate}
    \item \textbf{Supervised ViT-B} has scattered attention for both early and late TSC (see Figure \ref{fig:cradcam}). This suggests that the model fails to form semantically meaningful representations of defects under purely supervised training.
    \item \textbf{ViT-B pre-trained on ImageNet} shows sharp but misleading focus — attending to the boundary regions around solder pads even when defects are present. These regions are not causally linked to degradation. The model thus appears to exploit spurious correlations involving the general lifetime of certain LED types, compromising its generalizability to unseen defect modes.
    %\item  ViT-B self pre-trained on SAM data solely attends to actual defects such as cracks for later TSC with stronger degradation (Figure \ref{fig:cradcam} (b)) and attends to voids and the boundary regions between solder pads if no damages are visible (Figure \ref{fig:cradcam} (a)). This indicates that the model prioritizes actual defects and considers the general error-proneness of different LED types when no defects are visible.
    \item \textbf{ViT-B self pre-trained on SAM data}, in contrast, displays the behavior we expect from a  robust and generalizable model:
    \begin{itemize}
        \item For severe degradation cases (Figure \ref{fig:cradcam} (b)), attention is focused on real defects such as cracks.
        \item For defect-free samples (Figure \ref{fig:cradcam} (a)), attention shifts to error-prone areas such as void zones, or the bounding areas which indicate the type of the given LED, which is related to its general life expectancy.
    \end{itemize}
\end{enumerate}
\noindent Thus, our findings indicate that MAE self pre-training yields superior representations for defect detection compared to supervised training and pre-training on natural image data. This confirms our hypothesis that domain-specific pre-training can account for the  data requirements of transformers under the data sparsity in microelectronics.

\section{Discussion and Conclusion} In this paper, we employ vision transformers self pre-trained using masked-autoencoders for defect detection in microelectronics. Our methodology leverages the strong predictive capabilities of transformers while adapting to our specific target domain despite limited labeled data. For defect detection on a small dataset of microscale solder joints of high-power
LEDs, our approach demonstrates superior performance compared to various state-of-the-art CNN-based architectures and other transformer baselines. 

%We show that MAE self pre-training allows for the efficient application of pure transformer models to defect detection in microelectronics manufacturing, despite limited labelled data in the domain. For an SAM dataset of high-power LED solder joints labelled using TTA, our approach outperforms various state-of-the-art convolutional architectures for defect detection, achieving a substantial performance increase compared to the best evaluated baseline. This indicates that MAE self pre-training provides a practical solution for applying transformer models to defect detection in real-world inline quality control, offering fault-specific representations despite data limitations. 

While our MAE-based self pre-training framework demonstrates strong performance in defect detection, several directions offer potential for further enhancement. Future work could explore multi-modal pre-training that integrates additional signals such as thermal profiles or X-ray imaging to enrich learned representations. Combining images of multiple timesteps during pre-training may enable earlier and more accurate failure forecasting. Moreover, adaptive or defect-aware masking strategies could guide the model toward more semantically relevant features. Extending the framework to other manufacturing domains or unseen microelectronic devices would also test its robustness and generalizability. Finally, integrating our models into edge-computing environments for real-time, in-situ quality control remains a critical step toward practical deployment.

\begin{credits}

\subsubsection{\ackname} The presented work is part of the project MaWis-KI, which is supported by the Bavarian Ministry of Economic Affairs, Regional Development and Energy in the context of the Bavarian funding program for research and development "Information and Communication Technology".

\subsubsection{\discintname}
The authors have no competing interests to declare that are relevant to the content of this article.
\end{credits}
%
% ---- Bibliography ----
%
% BibTeX users should specify bibliography style 'splncs04'.
% References will then be sorted and formatted in the correct style.
%
\bibliographystyle{splncs04}
\bibliography{mybibliography}
%% Note that this preceding line implies that you store your BibTeX references in a file called 'mybibliography.bib'. If you instead store your references in a file with a different name, for instance 'references.bib', the preceding line should read '\bibliography{references}'. Whatever you do, DO NOT put the file name extension .bib inside the \bibliography command; this will trip up LaTeX compilers. 
%
% If you do not want to use BibTeX, you can also type up the bibliography exactly as you see fit, using the following structure:
%\begin{thebibliography}{8}
% Note that this number 8 reserves an amount of space (equal to the natural width of the given number) for the label of your references; if you have more than 9 references, you will want to change this number to 18. If you have more than 19 references, this number is best changed to 88. If you have more than 99 references, I salute you.

\end{document}